\documentclass[letterpaper]{article} 
\usepackage{aaai2026}  
\usepackage{times}  
\usepackage{helvet}  
\usepackage{courier}  
\usepackage[hyphens]{url}  
\usepackage{graphicx} 
\usepackage{natbib}  
\usepackage{caption} 
\usepackage{algorithm}
\usepackage{algorithmic}
\usepackage{amsfonts} 
\usepackage{amsmath}
\usepackage{booktabs}  
\usepackage{multirow}
\usepackage{subcaption}
\usepackage{cleveref}
\usepackage{bm}
\usepackage{colortbl}
\usepackage{xcolor}
\usepackage{enumitem}
\newcommand{\bestcell}[1]{\textbf{#1}}
\newcommand{\secondcell}[1]{\underline{#1}}
\usepackage{newfloat}
\usepackage{listings}
\usepackage{newfloat}
\usepackage{listings}
\DeclareCaptionStyle{ruled}{labelfont=normalfont,labelsep=colon,strut=off} 
\floatstyle{ruled}
\newfloat{listing}{tb}{lst}{}
\floatname{listing}{Listing}
\title{Mitigating Error Accumulation in Co-Speech Motion Generation via Global Rotation Diffusion and Multi-Level Constraints}
\author{
    Xiangyue Zhang\textsuperscript{\rm 1} \equalcontrib ,
    Jianfang Li \equalcontrib \textsuperscript{\rm 1}\thanks{Corresponding author.},
    Jianqiang Ren \textsuperscript{\rm 1} ,
    Jiaxu Zhang \textsuperscript{\rm 2}
}
\affiliations{
    \textsuperscript{\rm 1} Tongyi Lab, Alibaba Group \\
    \textsuperscript{\rm 2} Nanyang Technological University
}

\usepackage{bibentry}

\begin{document}

\maketitle

\begin{abstract}
Reliable long-horizon co-speech gesture generation requires precise motion representation and consistent structural priors across all joints. Existing generative methods typically operate on local joint rotations, which are defined hierarchically based on the skeleton structure. This leads to cumulative errors during generation, manifesting as unstable and implausible motions at end-effectors. In this work, we propose GlobalDiff, a diffusion-based framework that operates directly in the space of global joint rotations for the first time, fundamentally decoupling each joint’s prediction from upstream dependencies and alleviating hierarchical error accumulation. To compensate for the absence of structural priors in global rotation space, we introduce a multi-level constraint scheme. Specifically, a joint structure constraint introduces virtual anchor points around each joint to better capture fine-grained orientation. A skeleton structure constraint enforces angular consistency across bones to maintain structural integrity. A temporal structure constraint utilizes a multi-scale variational encoder to align the generated motion with ground-truth temporal patterns. These constraints jointly regularize the global diffusion process and reinforce structural awareness. Extensive evaluations on standard co-speech benchmarks show that GlobalDiff generates smooth and accurate motions, improving the performance by 46.0\% compared to the current SOTA under multiple speaker identities.
\end{abstract}

\begin{links}
\link{Project}{https://xiangyuezhang.com/GlobalDiff/}
\end{links}
\section{Introduction}
Holistic co-speech motion generation, which synchronizes body posture \cite{chhatre2024emotional,zhang2024speech}, hand gestures \cite{li2021audio2gestures,liu2022learning}, and facial expressions \cite{li2025wav2sem,peng2023emotalk} with speech, is an increasing focus in the field of computer vision. It plays a vital role in enabling virtual characters to communicate naturally and convincingly by modeling the full spectrum of non-verbal cues. This capability facilitates their deployment in avatars, interactive games, live-streaming platforms, and human-robot collaboration.

\begin{figure}
    \centering
    \includegraphics[width=0.45\textwidth]{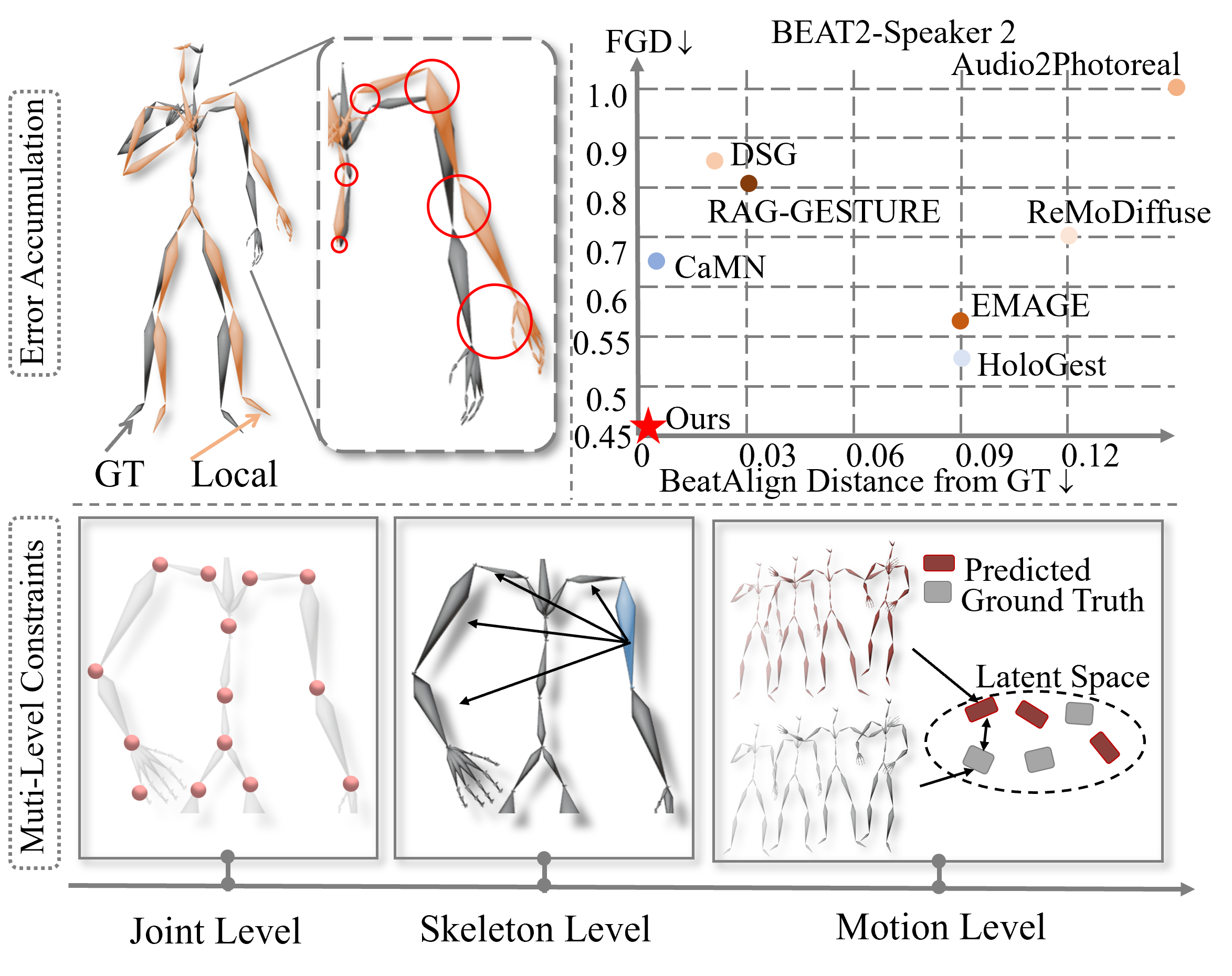}
    \caption{Overview of our motivation and solution. Local rotation diffusion leads to error accumulation in distal joints. Global rotation diffusion avoids this but lacks structural priors (top left). We address this with constraints at the joint, skeleton, and motion levels to ensure coherent and reasonable motion (bottom).}
    \label{fig:motivation}
\end{figure}

Recent advancements in generative modeling, particularly diffusion-based approaches \cite{zhu2023taming,alexanderson2023listen,yang2023diffusestylegesture,he2024co,chen2024enabling}, have notably improved the expressiveness and naturalness of generated gestures. Typically, these methods operate within a hierarchical local joint rotation space, wherein each joint's orientation is defined relative to its parent joint in a kinematic chain (e.g., SMPL-X \cite{pavlakos2019expressive}). While intuitive and structurally consistent, this local formulation inherently suffers from cumulative errors due to its recursive propagation mechanism. \textit{Minor inaccuracies in root or intermediate joint predictions can cause significant errors in the end-effectors, as shown in Figure \ref{fig:motivation}}, thereby substantially degrading the quality and stability of distal joints such as the hands, fingers, and feet—especially during prolonged and expressive motion sequences.

To address this limitation, we explore a new design direction: performing diffusion-based motion generation directly in the global joint rotation space, which we term GlobalDiff. By removing recursive dependencies among joints, global rotations fundamentally mitigate the problem of hierarchical error propagation, leading to more stable and coherent motion, even in long, fine-grained co-speech gestures.

However, this shift to global representation introduces a new challenge: the loss of natural structural constraints. Unlike local rotations, which implicitly preserve joint relationships through the hierarchical skeleton structure, global rotations treat each joint independently. Without additional guidance, this independence may lead to physically implausible poses or broken kinematic chains. Thus, while global rotation solves error accumulation, it also weakens structural consistency—a tradeoff that must be addressed to fully realize its benefits.

Motivated by these observations, we introduce a multi-level structure constraint designed to reinstate the lost structural coherence progressively. Inspired by human kinematic constraints, we incorporate constraints at three interconnected levels: joint-level, skeleton-level, and motion-level.

At the joint level, we propose to use a set of virtual anchor points attached to each joint to represent the joint-level structure. By aligning the transformed positions of these anchors—rotated by the predicted global rotation—with those derived from GT rotations, we resolve rotation ambiguity and achieve fine-grained supervision of joint orientation.

At the skeleton level, we propose to use an Angular Matrix (AM) computed from pairwise angles between all bone directions to represent the skeleton-level structure. By aligning the predicted AM with that from the GT motion, we constrain global bone relationships and enforce anatomically coherent and physically plausible skeleton configurations.

At the motion level, we propose to use temporal embeddings extracted via a multi-scale variational encoder \cite{kingma2013auto, ma2025followyourmotion,xiao2025recnet,yu2025mild} to represent the motion-level structure. By aligning the temporal features of the generated and ground-truth sequences, we ensure dynamic consistency, rhythm synchronization, and smooth temporal transitions throughout the motion.

These multi-level constraints and global rotation representation collectively guide the model to generate expressive, physically plausible motion from speech.

We validate GlobalDiff on several co-speech motion benchmarks. Experimental results show that it produces high-quality motion with improved structural consistency and expressiveness, outperforming existing approaches.

Our contributions are summarized as follows: 
\begin{itemize}[itemsep=0pt, parsep=0pt]
\sloppy 
\item We propose GlobalDiff, a diffusion-based co-speech motion generation framework that uses global joint rotations to resolve hierarchical error accumulation.

\item We design a multi-level structure constraint mechanism, consisting of joint-level virtual anchor points, skeleton-level bone consistency, and motion-level temporal coherence, progressively restoring structural plausibility.

\item We demonstrate that GlobalDiff achieves superior motion stability, anatomical correctness, and expressiveness compared to prior state-of-the-art methods on standard benchmarks.

\end{itemize}

\section{Related Work}
\textbf{Holistic Co-speech Motion Generation.}  
Co-speech motion generation aims to produce speech-aligned body movements. Early methods used GANs and diffusion models to improve realism and diversity \cite{habibie2021learning,ahuja2022low,ahuja2023continual,zhu2023taming,yang2023diffusestylegesture,zhi2023livelyspeaker}. Later work introduced semantic control through hierarchical designs or prompts \cite{qi2024emotiongesture,liu2022learning,liang2022seeg,chen2024enabling}. Recent approaches target holistic motion—jointly modeling face, hands, and torso—and fall into two main groups: VQ-VAE-based and diffusion-based.

Among VQ-VAE-based methods, TalkSHOW \cite{yi2023generating} handles face separately, while EMAGE \cite{liu2024emage} masks latent tokens and splits the body into four parts. ProbTalk \cite{liu2024towards} leverages PQ-VAE with decoding for rhythmic precision. SemTalk \cite{zhang2025semtalk} and EchoMask \cite{zhang2025echomask} adopt RVQ-VAE \cite{guo2024momask,lee2022autoregressive} with frame-level semantic focus. However, VQ-VAE-based methods suffer from the limited generation diversity. In this work, we use the diffusion model as our baseline to generate diverse motions.

\noindent \textbf{Diffusion-based Motion Generation.}
Diffusion models are widely used for human motion generation, including text-to-motion and co-speech tasks. DiffuseStyleGesture \cite{yang2023diffusestylegesture} first demonstrated expressive audio-conditioned gestures with rhythm and style control. Later works extended this direction: DiffSHEG \cite{chen2024diffsheg} modeled facial and body motions jointly; GestureDiffuCLIP \cite{ao2023gesturediffuclip} introduced CLIP-based style control; HoloGest \cite{cheng2025hologest} incorporated motion priors with part-wise diffusion; LivelySpeaker \cite{zhi2023livelyspeaker} used a two-stage semantic-then-audio pipeline; and RAG-Gesture \cite{mughal2025retrieving} added retrieval for stronger semantic alignment. Yet these methods rely on local joint rotations, where forward kinematics causes error growth along the kinematic chain. This leads to unstable end-effectors. We address this by predicting global joint rotations directly and applying joint-, skeleton-, and motion-level constraints to maintain structural consistency and expressive quality.

\begin{figure*}
    \centering
    \includegraphics[width=0.98\textwidth]{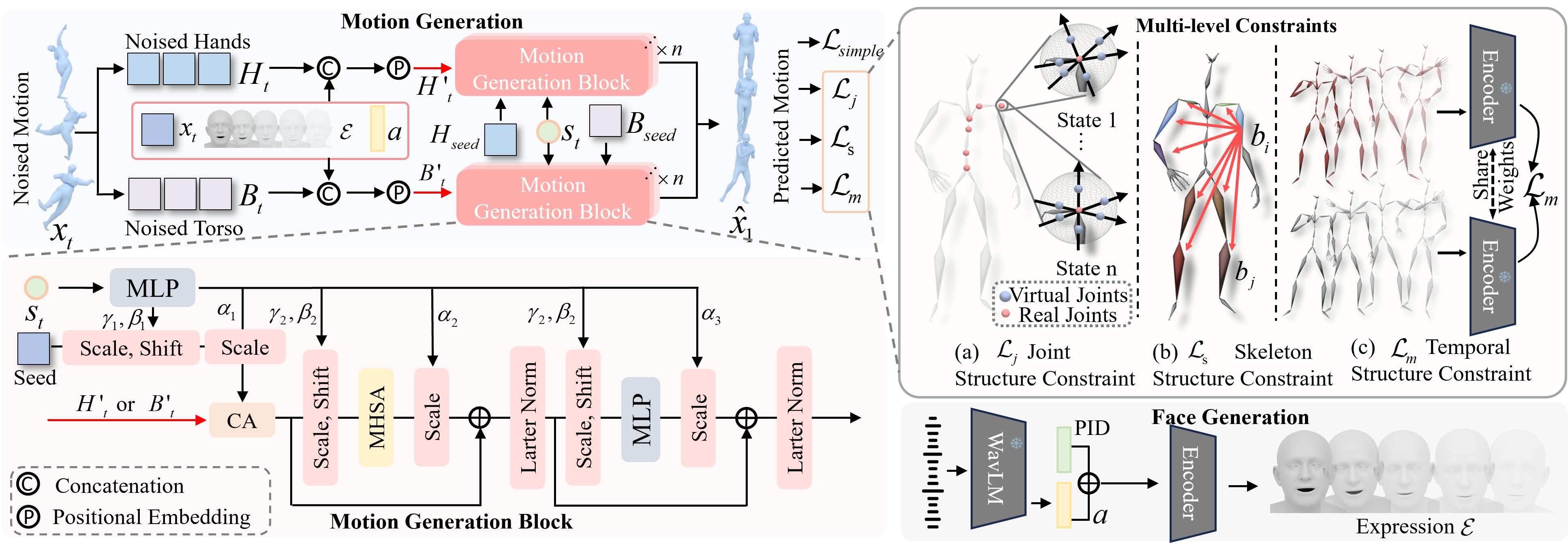}
    \caption{Overview of the GlobalDiff Framework. Our model generates consistent and expressive co-speech motion using the global rotation diffusion augmented with multi-level structural constraints. The diffusion model is conditioned on seed pose and prosodic features and predicts body motion through stacked motion generation blocks. To enforce structural plausibility, we introduce: (a) a Joint structure constraint using virtual anchor points to disambiguate orientations; (b) a Skeleton structure constraint that enforces angular consistency across adjacent bones by aligning the angular matrices; and (c) a Temporal structure constraint based on a shared multi-scale VAE encoder to preserve temporal dynamics. Facial expressions are generated in parallel from prosody using a transformer encoder.}

    \label{fig:method}
\end{figure*}

\section{Method}
\subsection{Preliminary}
\noindent \textbf{Local and Global Joint Rotations.} 
Based on the articulated skeletons, previous methods represent each joint's motion in its local coordinate frame. The local rotation \( R_k^{\text{local}} \in \text{SO}(3) \) defines the orientation of joint \( k \) relative to its parent in the skeletal hierarchy. In contrast, the global rotation \( R_k^{\text{global}} \in \text{SO}(3) \) expresses its absolute orientation in world coordinates. Typically, global rotations are recursively recovered by composing local rotations along the parent-child chain of the skeletal hierarchy:

\begin{equation}
R_k^{\text{global}} = R_{k-1}^{\text{global}} R_k^{\text{local}}, \quad R_1^{\text{global}} = R_1^{\text{local}},
\end{equation}

\noindent and the global position is:

\begin{equation}
q_k = R_{k-1}^{\text{global}} (t_k - t_{k-1}) + q_{k-1},
\end{equation}
where $t_k$ is the position of joint $k$ at rest state (T-pose).
This process, known as Forward Kinematics (FK), has been widely studied and applied in \cite{kucuk2006robot,aberman2020skeleton,zhangtapmo}. By unrolling this recursion, we can express \( R_k^{\text{global}} \) and \( q_k \) as cumulative products and sums:

\begin{equation}
R_k^{\text{global}} = R_1^{\text{local}} R_2^{\text{local}} \cdots R_k^{\text{local}},
\end{equation}
\begin{align}
q_k =\ 
&\left( \prod_{i=1}^{k-1} R_i^{\text{local}} \right) (t_k - t_{k-1}) \notag \\
+&\left( \prod_{i=1}^{k-2} R_i^{\text{local}} \right) (t_{k-1} - t_{k-2}) \notag \\
+&\cdots + q_1.
\end{align}

This shows that the global position of a joint depends on a chain of matrix multiplications and additions through all its ancestors in the skeleton. The deeper the joint is in the kinematic tree, the more transformations are involved.

\noindent \textbf{Error Accumulation in Local Methods.}  
Existing diffusion-based methods supervise the generated motion using positional losses (e.g., L2 distance between joint or vertex positions). To compute positions, they first predict local joint rotations \( R_k^{\text{local}} \) and apply Forward Kinematics.

However, FK recursively composes transformations along the kinematic chain, as in equation (4). As a result, even small errors in earlier local rotations (e.g., \( R_i^{\text{local}} \)) are multiplied and propagated forward, resulting in increasingly large deviations in joint positions at greater depths in the hierarchy, as shown in Figure \ref{fig:motivation}. This effect is especially severe at distal joints such as hands and feet, where accumulated errors lead to unstable or anatomically implausible motion—a phenomenon we term \textit{hierarchical error accumulation}. Moreover, backpropagation through the FK chain involves deep, as in equation 4, nonlinear transformations, causing gradient instability and hindering effective training.

\noindent \textbf{Our Global Rotation Prediction.}  
To avoid the recursive composition of local rotations, we directly predict each joint’s global rotation \( R_k^{\text{global}} \). All joints are therefore defined in a shared world frame, eliminating depth-dependent multiplication of transformations. Following the explicit path-sum form derived in the previous subsection, the global position of joint \( k \) is written as
\begin{equation}
q_k  
= q_{\text{root}} 
+ \sum_{(i \rightarrow j)\,\in\,\pi(k)} 
  R_i^{\text{global}} (t_j - t_i),
\end{equation}
where \( \pi(k) \) denotes the unique parent-child path from the root to joint \( k \), and \( t_j - t_i \) is the rest-pose offset between consecutive joints along this path. This form matches the unfolded FK formulation but replaces local rotations with predicted global rotations.

Because this position computation is additive along \(\pi(k)\) and avoids recursive rotation composition, each joint receives a direct and stable gradient with respect to its global rotation. This removes hierarchical error accumulation and improves robustness when training with positional supervision. Although direct global rotation prediction reduces the natural structural coupling enforced by standard FK, we introduce multi-level structure constraints in Section~\ref{sec:constrains} to restore these geometric relations.

\subsection{Overall Structure}

As shown in Figure \ref{fig:method}, our model operates under the conditional flow matching (CFM) framework and takes as input a noised motion sequence \( x_t \), audio features \( a \), speaker identity, and a short seed motion clip. It predicts the clean global joint rotations and translations \( x_1 \in \mathbb{R}^{T \times (J \times 6 + 3) } \), representing the full-body motion in 6D rotation format.

\noindent\textbf{Audio and Expression Conditioning.} Given a raw audio waveform, we extract high-level acoustic features \( a \in \mathbb{R}^{T \times d_a} \) using a pretrained WavLM encoder \cite{chen2022wavlm}. These features are then combined with the speaker identity vector PID and passed through a shallow Transformer encoder to directly estimate facial expressions \( \mathcal{E} \in \mathbb{R}^{T \times d_e} \) from the audio and PID embedding. This design is motivated by the near one-to-one correspondence between phoneme sequences and lip movements, allowing a lightweight yet accurate mapping for expression synthesis.

\noindent\textbf{Region-wise Motion Decomposition.} To enhance learning capacity, we divide the motion sequence into hand and torso components. The noised global pose input \( x_t \in \mathbb{R}^{T \times (J \times 6 + 3)} \) is decomposed into hand joints \( \mathbf{H}_t \) and torso joints \( \mathbf{B}_t \) based on a predefined joint partition. We then concatenate each of these with the expression and audio features along the channel dimension:
\begin{equation}
\mathbf{H}'_t = \text{Concat}(x_t, \mathbf{H}_t, \mathcal{E}, a), \quad 
\mathbf{B}'_t = \text{Concat}(x_t, \mathbf{B}_t, \mathcal{E}, a),
\end{equation}
yielding input embeddings enriched with multi-modal cues.

\noindent\textbf{Motion Generation Blocks.} 
The refined hand and torso features, \( \mathbf{H}'_t \) and \( \mathbf{B}'_t \), are passed through separate stacks of Motion Generation Blocks (MGBs), each built as a residual Transformer layer inspired by DiT \cite{peebles2023scalable}. Specifically, we apply FiLM-style \cite{perez2018film} affine transformations at multiple layers, conditioning on the style vector \( s_t = \text{Concat}(\text{MLP}(\text{PID}), \text{MLP}(t)) \), the flow step \( t \), and the seed motion. Each block also employs cross-attention to integrate style-aware context via a learned key-query mechanism.

\noindent\textbf{Flow Matching Objective.} 
Following \cite{tevet2022human}, our model adopts a simplified flow-matching formulation that directly learns to predict the clean motion sample \( x_1 \sim p_1 \) from an intermediate sample \( x_t = (1-t)x_0+tx_1 \), rather than estimating the continuous velocity field along the flow path. Given \( x_t \), the model is trained to output the corresponding target \( x_1 \) using a simple regression loss. The conditioning signal \( c \) consists of the expression features \( \mathcal{E} \), high-level audio feature \( a \), and the initial seed motion. The flow-matching loss is defined as:

\begin{equation}
\mathcal{L}_{\text{simple}} = \mathbb{E}_{t, x_0 \sim p_0, x_1 \sim p_1} \left\| f_\theta(x_t, c) - x_1 \right\|^2,
\end{equation}

\noindent where \( f_\theta \) is the model’s prediction network, and \( x_t \) is obtained by linearly interpolating between \( x_0 \) and \( x_1 \).

\subsection{Multi-Level Constraints}
\label{sec:constrains}
While our global rotation prediction removes error accumulation from recursive kinematic chains, it also discards the hierarchical structure inherent in local representation. As a result, directly learning global rotations can lead to physically implausible or unstable motions. To mitigate this, we introduce multi-level structure constraints at the joint, skeleton, and motion levels to restore structural coherence.

\noindent\textbf{Joint Structure Constraint.}  
Although our method learns global joint rotations through flow matching objectives, this rotation-based supervision alone does not provide joint structure information, e.g., skeleton length. To incorporate geometric information, MDM \cite{tevet2022human} suggests adding spatial position supervision of the joint \(\mathcal{L}_{pos}\). However, \(\mathcal{L}_{pos}\) does not sufficiently constrain the relationship between rotation and position, because simply constraining the position of the joint will cause multiple valid rotations to produce the same joint position, as shown in \ref{fig:method}, especially for the distal nodes. As can be seen from Equation (2), the position of the end node does not involve its own rotation. Therefore, we introduce joint-level constraints using virtual anchor points to introduce geometric priors into the learning process. These virtual points not only provide explicit rotation guidance, but also implicitly encode bone length and spatial structure, helping the model learn meaningful relationships between joint rotations and bone geometry.

To this end, besides \(\mathcal{L}_{pos}\), we introduce a joint structure constraint based on \emph{virtual anchor points}. For each joint \( k \), we define \( N \) non-coplanar points \( \{v_k^n\}_{n=1}^N \) in its local frame. Under predicted rotation \( R_k^{\text{global}} \), they are transformed to:
\begin{equation}
\hat{v}_k^n = R_k^{\text{global}} \cdot v_k^n,
\end{equation}
and compared with ground-truth rotated anchors:
\begin{equation}
\tilde{v}_k^n = R_k^{\text{gt}} \cdot v_k^n.
\end{equation}
Since the anchors span 3D space, matching \( \hat{v}_k^n \) to \( \tilde{v}_k^n \) uniquely constrains rotation.

We define the supervision loss as:
\begin{equation}
\mathcal{L}_j = \frac{1}{KN} \sum_{k=1}^{K} \sum_{n=1}^{N} \left\| \hat{v}_k^n - \tilde{v}_k^n \right\|_2^2.
\end{equation}

This constraint injects geometric priors into training and stabilizes rotation prediction, especially in expressive regions like hands and elbows. Although distal joints are more prone to rotational ambiguity, we apply this constraint to all joints to regularize global rotation learning across the entire skeleton and improve anatomical coherence.

\noindent\textbf{Skeleton Structure Constraint}.
While the joint-level constraint enforces local rotational fidelity, it alone cannot capture the global anatomical structure of the human skeleton. In particular, human motion is governed by interdependent skeletal geometry, where the orientation of one bone is implicitly constrained by the orientations of all others. Prior works that supervise each joint independently fail to model this global coordination, leading to implausible artifacts such as asymmetric bending, unnatural twisting, and broken kinematic continuity.
\begin{figure*}
    \centering
    \includegraphics[width=0.9\textwidth]{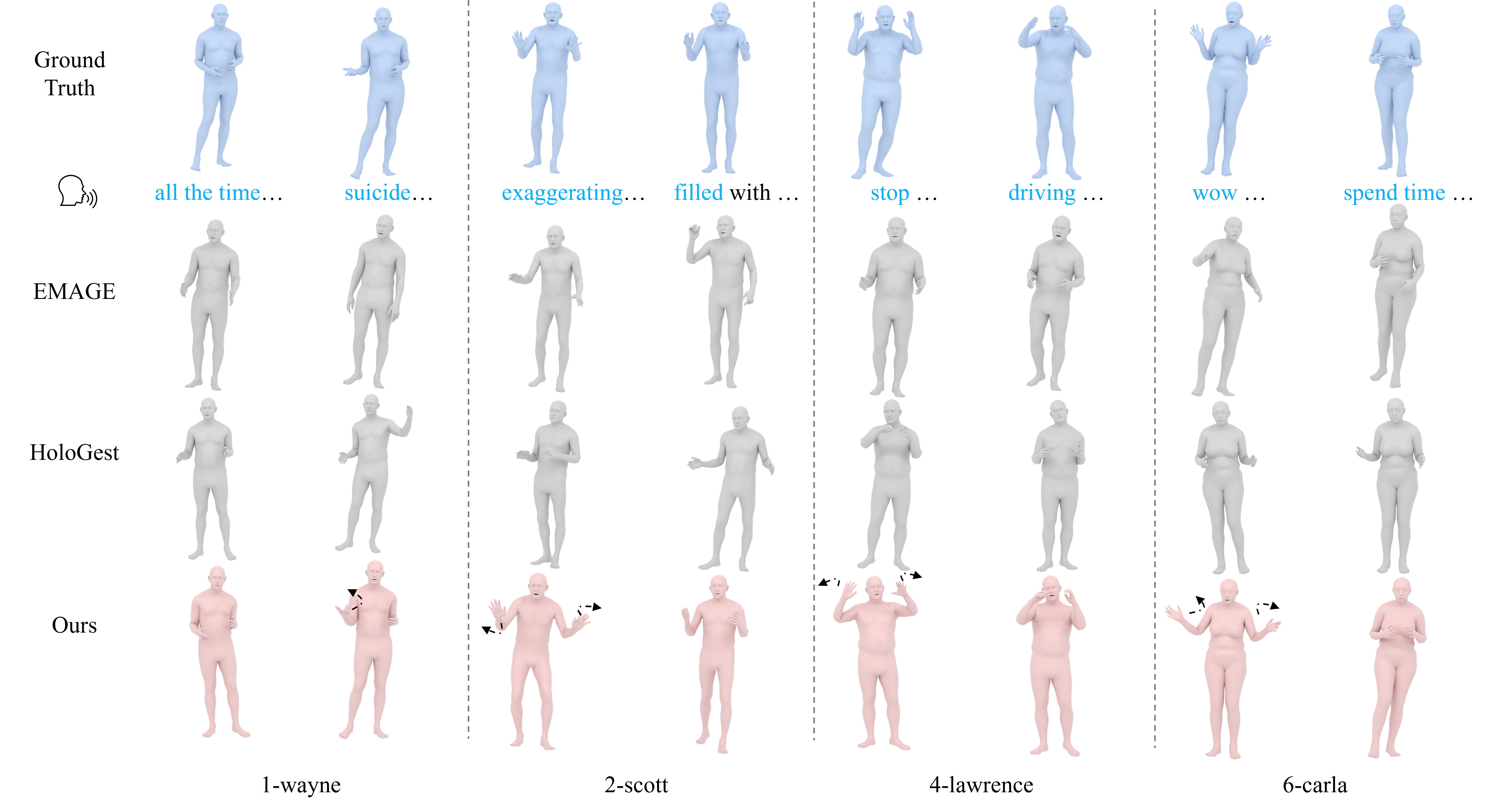}
    \caption{Visual comparison. Our GlobalDiff produces semantically meaningful gestures that align closely with the spoken content and speaker identity. For example, for the phrase “stop,” Our GlobalDiff generates symmetric and contextually appropriate hand gestures near the head, conveying a clear intent. In contrast, HoloGest and EMAGE often result in unbalanced or semantically incoherent motions, such as asymmetric arms or ambiguous limb orientations.}
    \label{fig:compare}
\end{figure*}

To model this high-level structural regularity, we propose a skeleton-level constraint based on a pairwise Angular Matrix (AM) that captures angular relations between all bone pairs. For each bone defined by a joint pair \( (k, j) \) with valid positional data, we define the unit bone direction vector in global space as:
\begin{equation}
b_{k \rightarrow j} = \frac{q_j - q_k}{\| q_j - q_k \|_2},
\end{equation}
where \( q_k \) and \( q_j \) are the global positions of joints \( k \) and \( j \). 

Next, we define an Angular Matrix \( \mathcal{A} \in \mathbb{R}^{K \times K} \), where each entry represents the cosine similarity between bone \( b_{k \rightarrow j} \) and every other bone \( b_{k' \rightarrow j'} \):
\begin{equation}
\mathcal{A}_{kj,k'j'} = b_{k \rightarrow j}^\top b_{k' \rightarrow j'}.
\end{equation}
This matrix encodes all pairwise bone orientation relations, thus capturing long-range skeletal dependencies and relative angular consistency between any pair of limbs, even if they are not directly connected in the kinematic tree.

Given predicted joint positions \( \{q_k\} \) and ground-truth positions \( \{\tilde{q}_k\} \), we compute their corresponding angular matrices \( \mathcal{A} \) and \( \tilde{\mathcal{A}} \). The skeleton structure constraint is defined as the mean squared error over all valid bone pairs:
\begin{equation}
\mathcal{L}_s = \frac{1}{|\mathcal{B}|} \sum_{(k,j),(k',j') \in \mathcal{B}} \left\| \mathcal{A}_{kj,k'j'} - \tilde{\mathcal{A}}_{kj,k'j'} \right\|_2^2,
\end{equation}
where \( \mathcal{B} \) is the set of all valid bone pairs with well-defined direction vectors.

\noindent\textbf{Temporal Structure Constraint}. 
While joint-level and skeleton-level constraints ensure spatial plausibility within individual frames, they do not capture the temporal structure of co-speech motion, which is inherently rhythmic and synchronized with the speech. Without explicit modeling of temporal dynamics, generated motions may appear erratic, lack rhythmic coherence, or fail to align with prosodic patterns. To address this, we introduce a motion-level constraint that enforces temporal consistency by aligning the dynamics of the generated motion sequence with those of the ground truth. Specifically, both the predicted motion \( \hat{X} = \{ \hat{x}_t \}_{t=1}^T \) and the reference motion \( X = \{ x_t \}_{t=1}^T \) are encoded into temporal embeddings using a shared multi-scale VAE \( g(\cdot) \) to capture both short- and long-term dependencies.

We compute a direct perceptual alignment loss based on mean squared error (MSE) between the temporal embeddings:
\begin{equation}
z^{\text{gen}} = g(\hat{X}), \quad z^{\text{gt}} = g(X),
\end{equation}
\begin{equation}
\mathcal{L}_m = \left\| z^{\text{gen}} - z^{\text{gt}} \right\|_2^2.
\end{equation}
This formulation encourages the model to match global motion dynamics in latent space while maintaining simplicity and stability in training.

\section{Experiments}
\subsection{Experimental Setup}
\noindent\textbf{Datasets.} 
For training and evaluation, we adopt the BEAT2 dataset \cite{liu2024emage}, which provides approximately 60 hours of 3D motion data. It contains 1,762 dialogue sequences, each lasting around 65 seconds on average. While prior methods typically report results on a fixed subset of speakers, we evaluate our model on both the full test set and the subset corresponding to speaker "Scott" to support fair comparison with single-speaker setups and assess generalization across diverse speakers, as recommended by \cite{mughal2025retrieving}.
\begin{figure}
    \centering
    \includegraphics[width=0.44\textwidth]{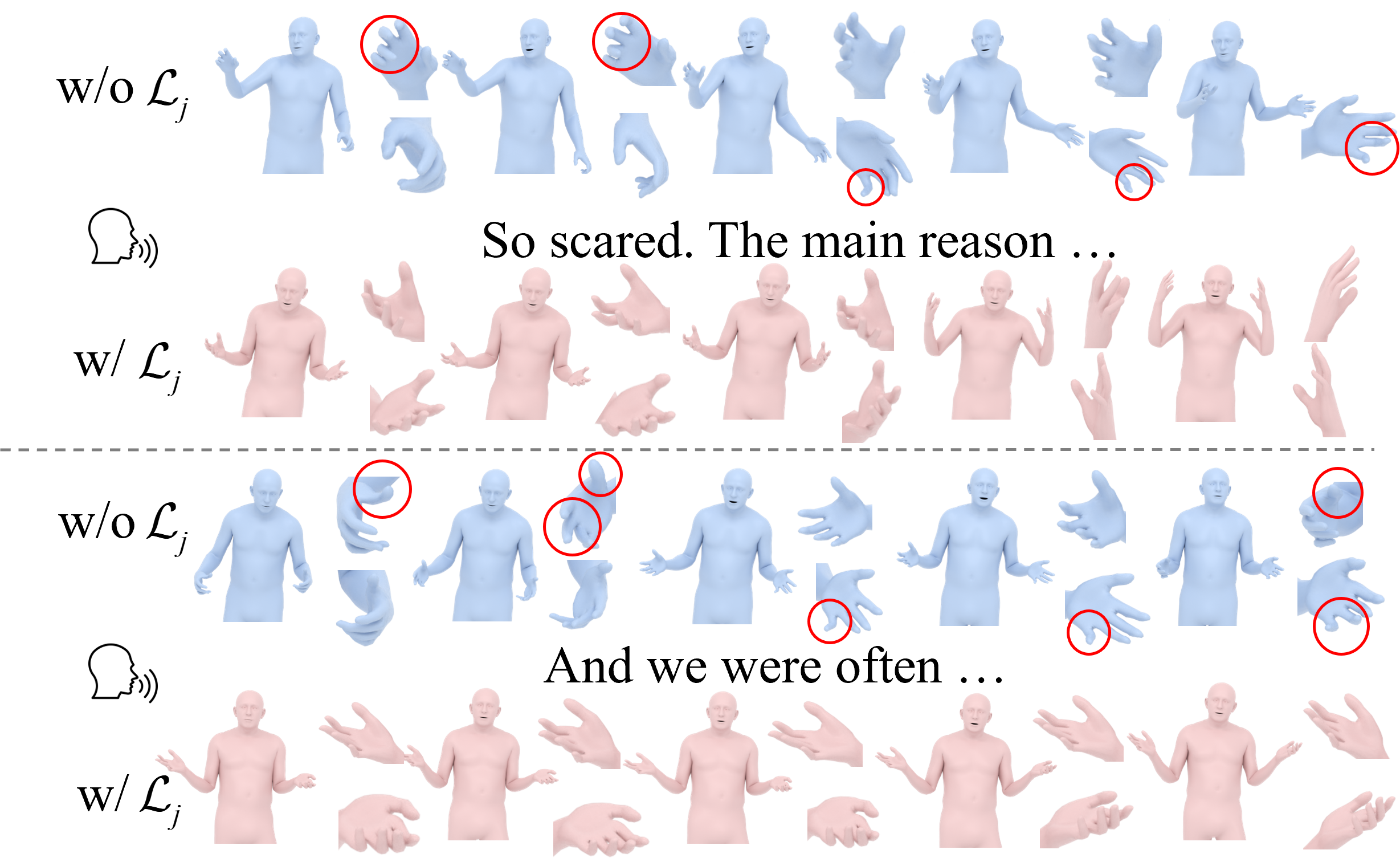}
    \caption{Qualitative study on the effect of  \(\mathcal{L}_j\). Without \(\mathcal{L}_j\), finger poses often appear anatomically implausible.}
    \label{fig:joint}
\end{figure}


\noindent\textbf{Implementation Details.}
Training is performed on four NVIDIA V100 GPUs for 1,000 epochs with a batch size of 128 and takes about 17 hours. We use the ADAM optimizer with a learning rate of 1e-4. The number of virtual nodes is set to 6. Following \cite{liu2024emage}, training begins with an 8-frame seed pose. For streaming, the last 8 frames of each clip serve as the seed for the next, so long sequences require only the initial 8 frames.

\noindent\textbf{Metrics.}
We evaluate motion using four criteria. Fréchet Gesture Distance (FGD) \cite{yoon2020speech} measures distribution-level realism. Diversity \cite{li2021audio2gestures} captures variation across generated clips through mean L1 differences. Beat Alignment (BeatAlign) \cite{li2021ai} assesses rhythm synchrony between speech and motion. Facial accuracy is measured with vertex MSE \cite{yang2023diffusestylegesture}.

\subsection{Qualitative Results}

\noindent\textbf{Qualitative Comparisons.}
Figure~\ref{fig:compare} shows that GlobalDiff produces semantically clear and physically stable co-speech motions across all speaker identities. We compare our model with EMAGE and HoloGest on four speakers—Wayne (ID1), Scott (ID2), Lawrence (ID4), and Carla (ID6). RAG-GESTURE \cite{mughal2025retrieving} is omitted due to unavailable public results. For phrases such as \textit{“exaggerating”} and \textit{“wow”}, GlobalDiff generates wide, well-formed arm and hand motions that reflect emphasis and surprise, whereas EMAGE tends to restrict movement and HoloGest often produces flat poses. For \textit{“driving”}, our model synthesizes steering-like motions, while the baselines revert to generic gestures. For \textit{“all the time”}, GlobalDiff maintains stable chest-level motions across speakers, while EMAGE introduces imbalance and HoloGest reduces motion detail. These observations highlight the value of global rotation prediction and our structural constraints.

\begin{figure}
    \centering
    \includegraphics[width=0.44\textwidth]{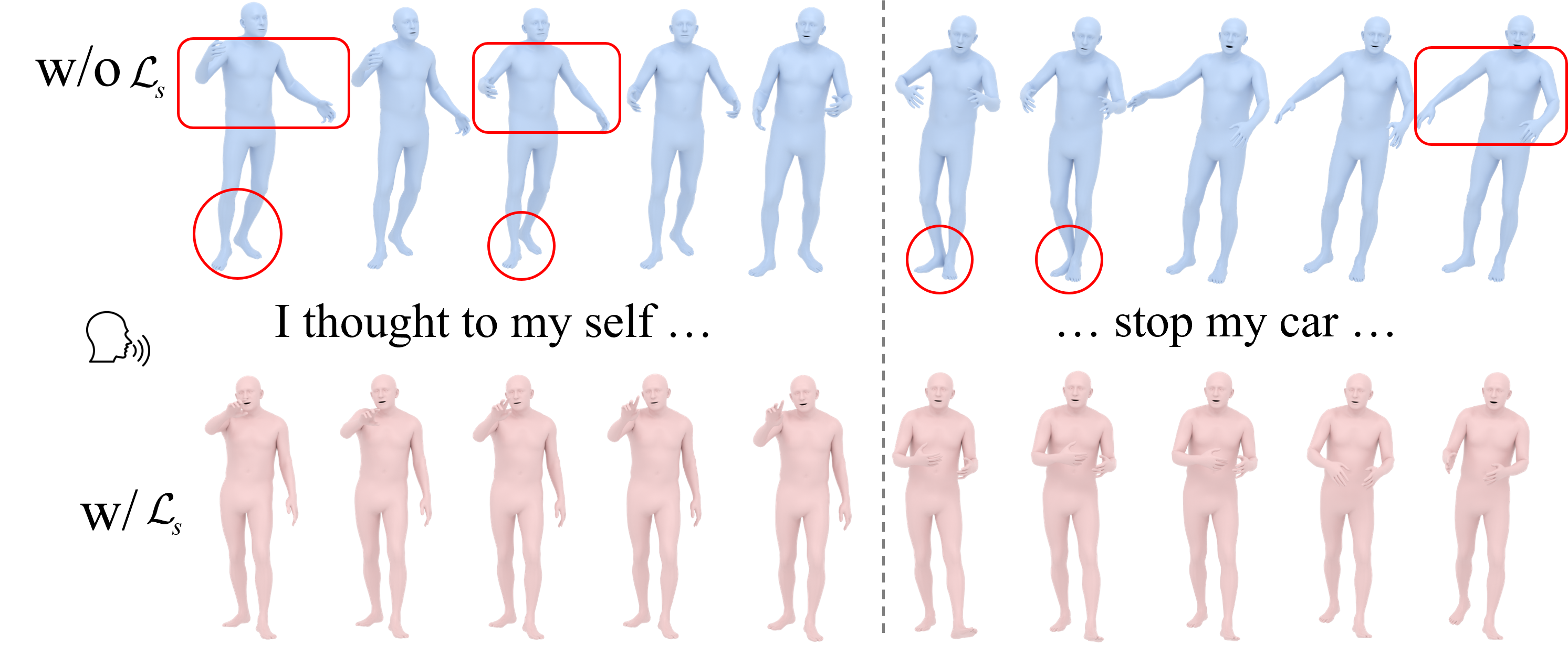}
    \caption{Qualitative study on the effect of  \(\mathcal{L}_s\). Without \(\mathcal{L}_s\), motion becomes structurally incoherent, exhibiting unbalanced posture and unstable foot placement.}
    \label{fig:skeleton}
\end{figure}

\begin{figure}
    \centering
    \includegraphics[width=0.44\textwidth]{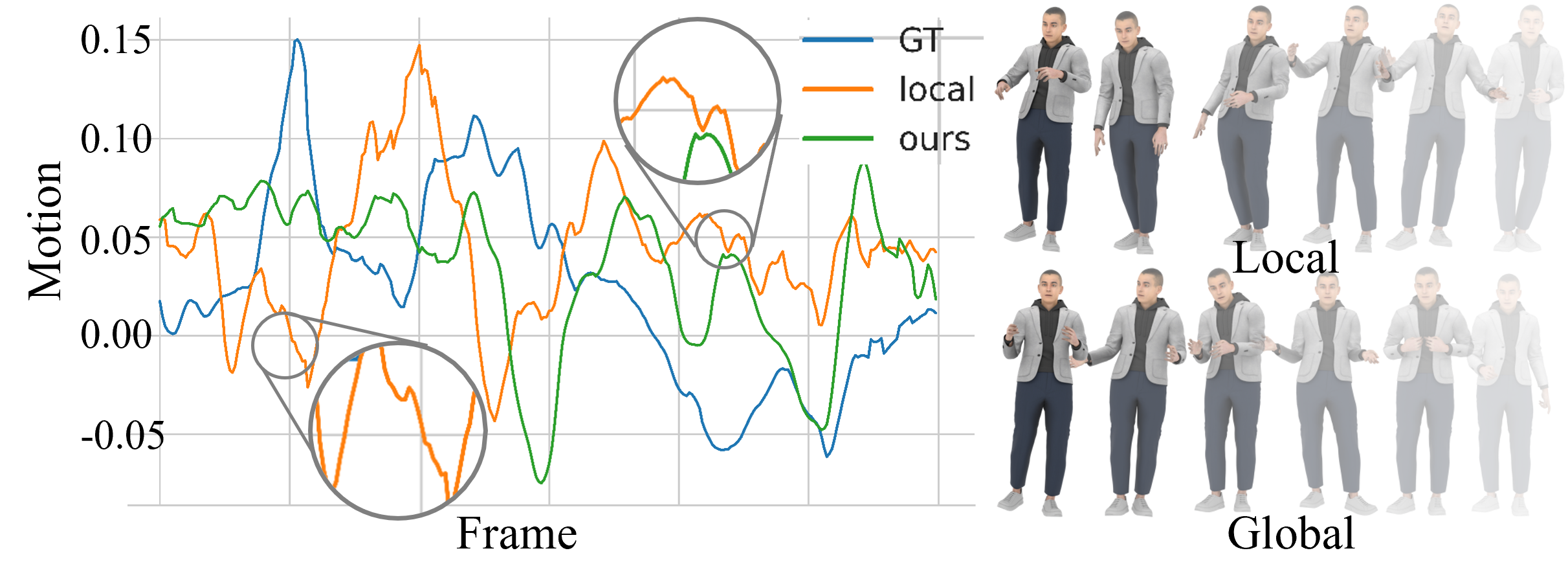}
    \caption{Ablation Study on Global vs. Local Rotation Prediction. Our global method produces smoother fingertip motion and more stable body posture compared to the noisy and distorted results from local rotation prediction. }
    \label{fig:distal_ablation}
\end{figure}

\begin{table*}
\centering
\setlength{\tabcolsep}{6pt}
\setlength{\fboxsep}{1pt}
\renewcommand{\arraystretch}{1.2}
\resizebox{0.9\textwidth}{!}{%
\begin{tabular}{lcccccccc}
\toprule
& \multicolumn{4}{c}{\textbf{1 Speaker}} & \multicolumn{4}{c}{\textbf{All Speakers}} \\
\cmidrule(lr){2-5} \cmidrule(lr){6-9}
\textbf{} & FGD$\downarrow$ & BeatAlign$\rightarrow$ & Diversity$\rightarrow$ & MSE$\downarrow$ 
         & FGD$\downarrow$ & BeatAlign$\rightarrow$ & Diversity$\rightarrow$ & MSE$\downarrow$ \\
\cmidrule(lr){2-5} \cmidrule(lr){6-9}
GT & -- & 0.703 & 11.97 & -- & -- & 0.477 & 7.29 & -- \\
\cmidrule(lr){1-9}
\rowcolor{white} CaMN \cite{liu2022beat} & 0.604 & \secondcell{0.711} & 9.97 & -- 
                                  & 0.512 & 0.200 & 5.58 & -- \\
\rowcolor{white} Audio2Photoreal \cite{ng2024audio2photoreal} & 1.02 & 0.550 & 12.47 & --
                                            & 0.849 & 0.326 & 6.24 & -- \\
\rowcolor{white} ReMoDiffuse \cite{zhang2023remodiffuse} & 0.702 & 0.824 & \secondcell{12.46} & --
                                         & 1.120 & 0.218 & 5.06 & -- \\
\rowcolor{white} DSG \cite{yang2023diffusestylegesture} & 0.881 & 0.724 & 11.49 & --  
                                   & 1.174 & 0.734 & 11.12 & -- \\
\rowcolor{white} HoloGest \cite{cheng2025hologest} & \secondcell{0.534} & 0.795 & 14.15 & --  
                                   & 0.646 & 0.803 & 13.53 & -- \\
\rowcolor{white} EMAGE \cite{liu2024emage} & 0.570 & 0.793 & 11.41 & 7.680  
                                   & 0.692 & 0.284 & 6.06 & 6.908 \\
\rowcolor{white} RAG-GESTURE \cite{mughal2025retrieving} & 0.808 & 0.734 & \bestcell{11.97} & --
                                              & \secondcell{0.487} & \bestcell{0.514} & \secondcell{9.94} & --\\
\cmidrule(lr){1-9}                                        
\rowcolor{white} \textbf{Ours} & \bestcell{0.478} & \bestcell{0.705} & 13.73 & \secondcell{6.330} & \bestcell{0.263} & \secondcell{0.404} & \bestcell{8.24} & \bestcell{4.144} \\
\bottomrule
\end{tabular}}
\caption{Comparison with state-of-the-art methods trained on BEAT2. We report MSE\(\times10^{-8}\) for simplicity}
\label{tab:sota}
\end{table*}

\begin{table}[t]
    \centering
    \resizebox{0.48\textwidth}{!}{%
    \begin{tabular}{lccc}
        \toprule
        \textbf{Method} & \textbf{FGD$\downarrow$} & \textbf{BeatAlign$\rightarrow$} & \textbf{Diversity$\rightarrow$}\\
        \midrule
        GT & -- & 0.703 & 11.97\\
        \cmidrule(lr){1-4}
        Ours(local) & 0.594 & 0.578 & 9.33 \\
        Ours(global) & 0.592 & 0.693 & 13.08  \\
        + \(\mathcal{L}_{j}\) & 0.574 & 0.665 & 12.30   \\
        + \(\mathcal{L}_{j}\) + \(\mathcal{L}_{s}\) & 0.517 & 0.593 & 13.78  \\
        + \(\mathcal{L}_{j}\) + \(\mathcal{L}_{s}\) + \(\mathcal{L}_{m}\) & \textbf{0.478} & \textbf{0.705} & 13.73 \\
        \bottomrule
    \end{tabular}}
    \caption{Ablation study on the effectiveness of each component in GlobalDiff on speaker 2}
    \label{tab:compare1}
\end{table}

\noindent\textbf{Joint Structure Constraint.}
We visualize hand close-ups from motions generated with and without \(\mathcal{L}_j\). Without \(\mathcal{L}_j\), the model frequently produces anatomically implausible joint rotations—such as unnaturally flipped thumbs or twisted little fingers (highlighted in red)—due to the under-constrained nature of rotation prediction. In contrast, incorporating \(\mathcal{L}_j\) guides the model to produce structurally consistent and physically realistic finger orientations by supervising global rotations via virtual anchor points.

\noindent\textbf{Skeleton Structure Constraint.}
As shown in Figure~\ref{fig:skeleton}, without \(\mathcal{L}_s\), the generated motions exhibit structural incoherence, such as unnatural leaning, unbalanced steps, and asymmetric limb configurations. These artifacts arise from the lack of explicit constraints on angular relationships between bones. In contrast, incorporating \(\mathcal{L}_s\) enforces consistent bone orientations across the body, resulting in smoother and more anatomically coherent movements.

\noindent\textbf{Ablation Study on Global vs. Local Rotation Prediction.}  
Figure~\ref{fig:distal_ablation} presents a qualitative comparison between our global rotation strategy, traditional local rotation methods, and ground truth (GT). On the left, we plot the trajectory of the right middle fingertip over 300 frames from the test sequence ``2-scott-0-103-103''. The local method suffers from high-frequency oscillations and instability, while our global method produces a smoother trajectory that aligns more closely with GT. On the right, snapshots sampled every 50 frames show that the local method introduces body tilt and severe finger deformation, especially at distal joints. In contrast, our method maintains natural body posture and stable articulation, demonstrating its effectiveness in mitigating hierarchical error accumulation.

\subsection{Quantitative Results}

\noindent\textbf{Comparison with Baselines.}
Table~\ref{tab:sota} provides a detailed quantitative comparison between our method and several state-of-the-art baselines on both single-speaker and multi-speaker co-speech motion generation. Our method consistently achieves the best performance across nearly all metrics. Specifically, we obtain the lowest FGD and comparable MSE, indicating superior spatial fidelity and motion reconstruction accuracy. On BeatAlign, our method performs comparably with the top baselines, reflecting robust temporal alignment with speech rhythm. Although RAG-GESTURE attains the closest diversity with GT on the single-speaker setting, our method maintains competitive diversity scores while preserving structural integrity and semantic consistency. These results highlight the strength of our global rotation modeling and multi-level structural constraints in producing accurate, expressive, and speaker-agnostic co-speech motion.

\noindent\textbf{Ablation Study on Components.}
Table~\ref{tab:compare1} presents the results of an ablation study evaluating the impact of each major component in our GlobalDiff pipeline. Starting from a local-rotation baseline with \(\mathcal{L}_{pos}\), we observe limited performance in both FGD and BeatAlign, highlighting the limitations of hierarchical propagation. Switching to global rotations improves rhythm alignment and motion diversity. Introducing $\mathcal{L}_j$ further enhances BeatAlign and diversity, indicating better semantic expressiveness and orientation precision. Adding $\mathcal{L}_s$ improves FGD substantially, reflecting enhanced anatomical plausibility. Finally, incorporating $\mathcal{L}_m$ brings all metrics to their best levels, demonstrating that our multi-level structure design jointly contributes to realism, rhythm alignment, and expressive richness.

\begin{figure}
    \centering
    \includegraphics[width=0.45\textwidth]{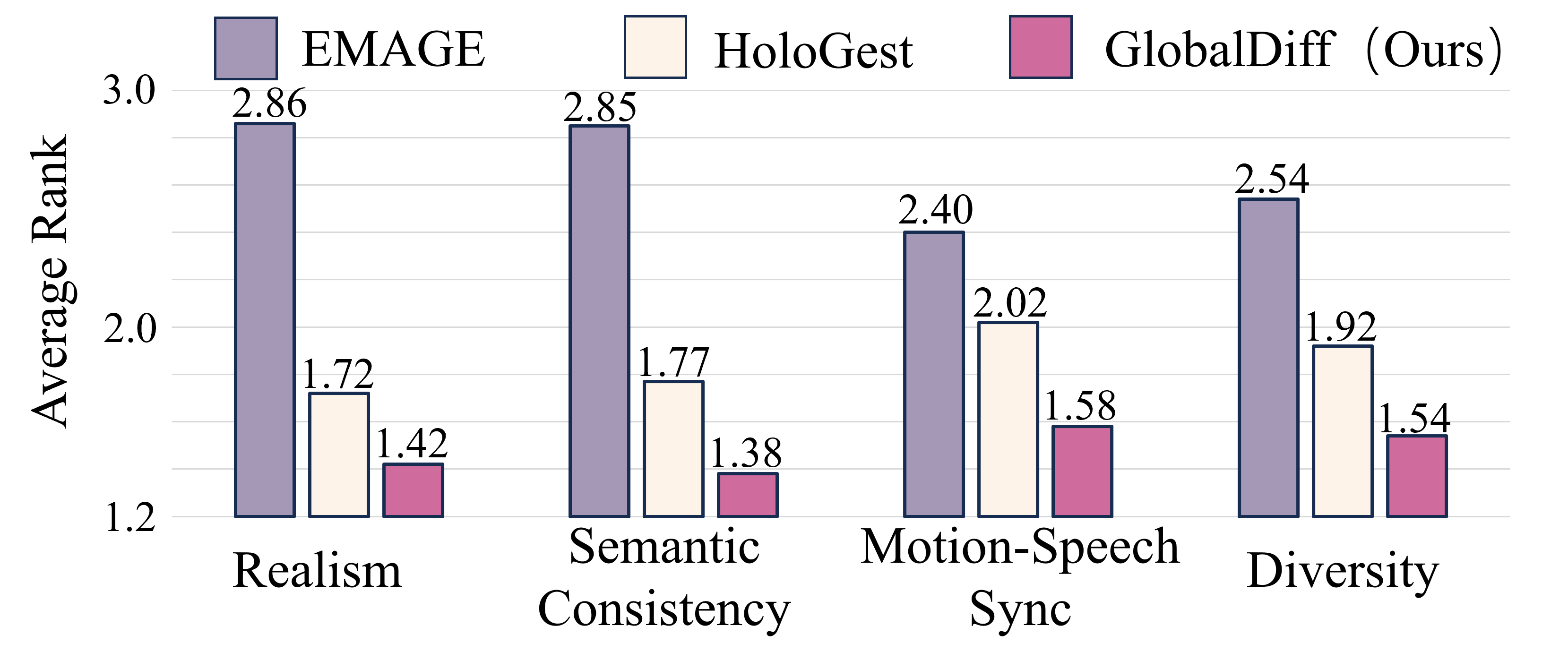}
    \caption{Results of the user study.} 
    \label{fig:user}
\end{figure}

\noindent \textbf{User Study.}
We conducted a user study with 10 video samples and 28 participants, evaluating realism, semantic consistency, and motion-speech synchrony. Participants were asked to rank anonymized and shuffled outputs from GlobalDiff, EMAGE, and HoloGest. As shown in Figure~\ref{fig:user}, GlobalDiff received higher preference across all metrics. 

\section{Conclusion}
In this work, we presented GlobalDiff, a novel diffusion-based framework for holistic co-speech motion generation that addresses the fundamental issue of hierarchical error accumulation inherent in local joint rotation approaches. By leveraging global joint rotations, GlobalDiff effectively decouples the prediction process across joints, significantly reducing cumulative errors, especially at distal limbs. To overcome the challenge of structural inconsistency arising from the loss of implicit hierarchical constraints, we introduced a progressive multi-level constraint scheme. Our joint-level constraint employs virtual anchor points for precise orientation guidance, the skeleton-level constraint ensures angular coherence among bones, and the motion-level constraint enforces temporal consistency through a multi-scale variational encoder. Extensive evaluations on standard co-speech benchmarks demonstrate that GlobalDiff not only achieves state-of-the-art performance in terms of motion stability, structural coherence, and expressiveness but also establishes a robust baseline for future research in structurally-aware global rotation diffusion methods.

\section*{Acknowledgments}
This work was supported by Alibaba Research Intern Program. The numerical computation was supported by Tongyi Lab, Alibaba Group


\end{document}